\documentclass{article}
\usepackage{ijcai26}

\usepackage{times}
\usepackage{soul}
\usepackage{url}
\usepackage[hidelinks]{hyperref}
\usepackage[utf8]{inputenc}
\usepackage[small]{caption}
\usepackage{graphicx}
\usepackage{amsmath}
\usepackage{amsthm}
\usepackage{booktabs}
\usepackage{algorithm}
\usepackage{algorithmic}
\usepackage[switch]{lineno}

\usepackage{amssymb}
\usepackage{tabularx}
\usepackage{multirow}
\usepackage{subcaption}

\title{DUALFloodGNN: Physics-informed Graph Neural Network for Operational Flood Modeling}

\author{
Carlo Malapad Acosta$^1$\and
Herath Mudiyanselage Viraj Vidura Herath$^2$\and
Jia Yu Lim$^1$\and
Abhishek Saha$^{3}$\and
Sanka Rasnayaka$^1$\And
Lucy Marshall$^2$\\
\affiliations
$^1$Department of Computer Science, School of Computing, National University of Singapore\\
$^2$School of Civil Engineering, Faculty of Engineering, The University of Sydney\\
$^3$Delft Institute of Applied Mathematics, Delft University of Technology\\
\emails
e1391136@u.nus.edu,
viraj.herath@sydney.edu.au
}

\begin{document}

\maketitle

\begin{abstract}
    Flood models inform strategic disaster management by simulating the spatiotemporal hydrodynamics of flooding. While physics-based numerical flood models are accurate, their substantial computational cost limits their use in operational settings where rapid predictions are essential. Models designed with graph neural networks (GNNs) provide both speed and accuracy while having the ability to process unstructured spatial domains. Given its flexible input and architecture, GNNs can be leveraged alongside physics-informed techniques with ease, significantly improving interpretability and generalizability. We introduce a novel flood GNN architecture, DUALFloodGNN, which embeds physical constraints at both global and local scales through explicit loss terms. The model jointly predicts water volume at nodes and flow along edges through a shared message-passing framework. To improve performance for autoregressive inference, model training is conducted with a multi-step loss  enhanced with dynamic curriculum learning. Compared with standard GNN architectures and state-of-the-art GNN flood models, DUALFloodGNN achieves substantial improvements in predicting multiple hydrologic variables (e.g., water volume, flow, and depth) while maintaining high computational efficiency. The model is open sourced at \url{https://github.com/acostacos/dual_flood_gnn}. The dataset is open sourced at \url{https://hdl.handle.net/2123/35293} with the DOI 10.25910/9xav-0s86.
\end{abstract}

\section{Introduction}

Floods pose a substantial socioeconomic threat to modern society, resulting in losses of more than 3 trillion USD in the past 25 years and affecting up to 1.81 billion individuals globally~\cite{Rentschler2022,UNDRR2020}. Furthermore, the frequency and severity of such events are only projected to increase due to climate change~\cite{Tabari2020}. Effective mitigation of these repercussions requires detailed flood information. In such cases, hydrodynamic flood models play a crucial role by simulating water behavior across space and time, enabling data-driven disaster management and flood-resilient engineering designs~\cite{Kumar2023}. Conventional physics-based numerical models capture flood dynamics with high accuracy by solving the governing equations of fluid flow. However, they require substantial computation time, limiting their effectiveness for real-time or operational use ~\cite{Sun2023}.

To address this, data-driven surrogate models forecast flooding through statistical correlations, resulting in much faster computation speed. In particular, deep learning (DL) has the ability to automatically learn and extract complex non-linear relationships from data, making it suitable for modeling flood movement. Research continues to assess the efficacy of various architectures including multilayer perceptrons (MLPs)~\cite{Ahmadlou2021,Xie2021}, recurrent neural networks (RNNs)~\cite{Fang2021,Rahimzad2021} and convolutional neural networks (CNNs)~\cite{Herath2025,Hosseiny2021,Neo2025}. Although suitable for real-time forecasting,  DL models are often criticized as "black boxes", due to a limited understanding of their predictive paradigm~\cite{Devi2015}. Such an opacity presents a significant disadvantage in physics-based disciplines, where explainability is essential to ensure consistency with physical laws.

\begin{figure}[t]
\centering
\includegraphics[width=\columnwidth]{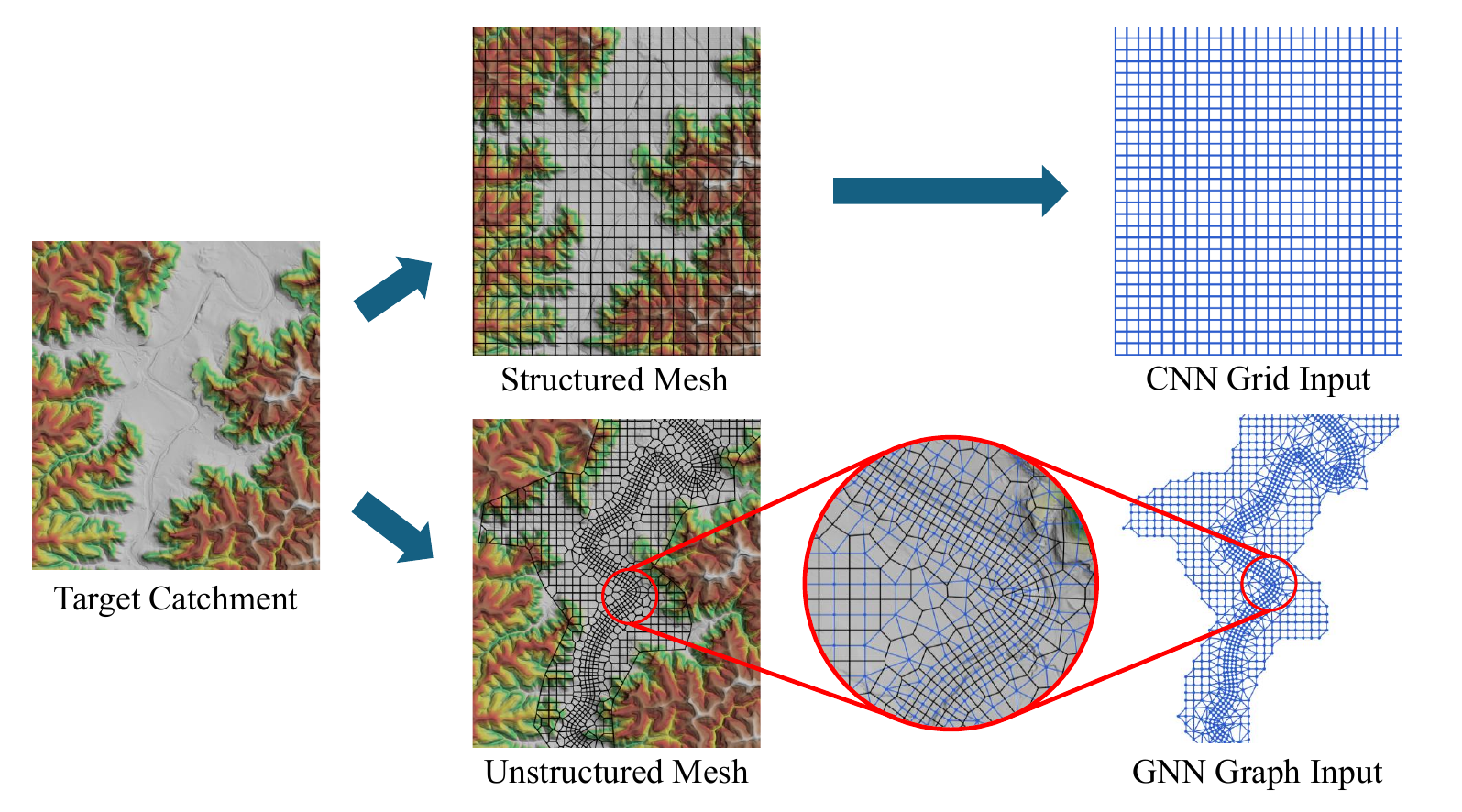}
\caption{Discretization of a target catchment. A structured mesh produces a uniform grid suitable for CNNs. A more flexible unstructured mesh can be translated into a graph for GNNs. The centroid of each mesh cell is represented as a node, while connections between adjacent cell centers define the edges as shown in the magnified section.}
\label{fig:discretization}
\vspace{-3mm}
\end{figure}

Physics-informed DL models combine the strengths of data-driven and physics-based approaches by embedding physical constraints into DL architectures, enabling them to learn the underlying behavior of physical systems ~\cite{Sharma2023}. The main challenge of this approach lies in the complexity of aligning existing algorithms in both domains. An example of such a case is illustrated in figure~\ref{fig:discretization}. Physics-based numerical approximations involve the discretization of the area into a mesh structure, which is typically integrated in DL models through CNNs, given their aptitude with spatial input~\cite{Gao2022}. However, this methodology is also fundamentally constrained by the nature of CNNs, which are restricted to processing structured grid meshes. This limits physical fidelity, as this discretization scheme cannot adequately represent the irregular topographies and non-uniform fluid dynamics characteristic of flooding scenarios. Graph neural networks (GNNs) were developed to directly address this limitation by generalizing CNN-based convolution for arbitrary spatial data represented as graphs~\cite{Bronstein2017}. This representation enables a more flexible description of hydrodynamic quantities through an unstructured mesh and has been shown to improve generalization in physical systems~\cite{Thangamuthu2022}.

Integration of physics-informed techniques with GNN flood models is still an emerging field of research. For example, HydroGraphNet~\cite{Taghizadeh2025} incorporates the principle of mass conservation at a global graph-level through a regularization term in the loss function. However, local mass conservation is only implicitly assumed, as global mass balance does not guarantee the same behavior in smaller regions. We address this research gap by incorporating an additional loss term explicitly promoting node-level mass balance. To enable this, we diverge from previous approaches~\cite{Bentivoglio2023,Bentivoglio2024} by mapping the time-varying water flow as an edge attribute, thus preserving its vector characteristics. This graph mapping provides more intuitive physical meaning, and closely mimics the capabilities of physics-based numerical models. However, this also introduces the challenge of predicting both node and edge features. Existing domain-agnostic GNNs created for this purpose~\cite{Jiang2019,Yang2020} require the generation of a line graph, which typically contains more nodes and edges, increasing computational overhead. We develop a novel architecture that performs simultaneous node and edge prediction given only the original graph utilizing two separate decoders. We coin our approach \textbf{D}ouble-target \textbf{U}niversally \textbf{A}nd \textbf{L}ocally constrained Flood GNN (\textbf{DUALFloodGNN}).

Our contributions can be summarized as follows:
\begin{itemize}
    \item We develop a physics-informed loss function that promotes mass conservation at both global and local scales through multiple regularization terms, preserving conservation within localized areas of the computational mesh and ensuring physically realistic flood dynamics.
    \item To facilitate the computation of local mass balance, we design an architecture capable of simultaneously predicting features for nodes (water volume) and edges (water flow). Joint modeling is performed through shared message generation, which utilizes their physical correlation while requiring only the original graph structure.
    \item To the best of our knowledge, this is the first study to benchmark performance across different GNN-based flood models. Experimental results validate the effectiveness of DUALFloodGNN, which outperforms both standard GNN baselines and state-of-the-art GNN flood models \cite{Bentivoglio2023,Taghizadeh2025}.
\end{itemize}

\section{Background and Related Work}

\paragraph{Graph Neural Networks.}

GNNs have found widespread use in the processing of graph-structured data~\cite{Hamilton2018,Kipf2017,Velickovic2018}. GNNs generalize CNN-based convolution for non-Euclidean domains~\cite{Bronstein2017} through the message passing framework~\cite{Gilmer2017}. Given a node $i$ in the graph, its node embedding $h_i^l$ for the current layer $l$ is updated to $h^{l+1}_i$ based on its local neighborhood $\mathcal{N}(i)$ through Eq. \ref{eq:msg_pass_def}:
\begin{equation}
    h^{(l+1)}_i =  U(h^{(l)}_i, A(M(h^{(l)}_i, h^{(l)}_j))); \: \forall j \in \mathcal{N}(i)
    \label{eq:msg_pass_def}
\end{equation}

\noindent
where $h_j$ are the neighbor embeddings and $M$, $A$ and $U$ are the message, aggregate and update functions. This is repeated for each node in the graph, creating an updated graph representation after one full cycle. Each layer in a GNN corresponds to one message passing iteration, which can be repeated to enable further message propagation.

\paragraph{Node and Edge Prediction.}

Convolution with edge-node switching network (CensNet) ~\cite{Jiang2019} learns both node- and edge-level features through alternating spectral graph convolutions using the original graph for node embeddings and the line graph for edge embeddings. Node and Edge Neural Network (NENN) ~\cite{Yang2020} extends this approach to spatial-based message passing and incorporates attention to both node- and edge-level layers.


\paragraph{GNNs for Flood Modeling.}

Initial studies adopted existing architectures for the hydrological context~\cite{Farahmand2023,Santos2023}. Further studies incorporated RNN components to explicitly capture the temporal aspect of flood dynamics~\cite{Kazadi2024,Roudbari2024,Zhao2020}, with some utilizing heterogeneous graph structures to encode diverse relationships within the system~\cite{Jiang2024,Luo2024}.

\paragraph{Physics-informed Flood GNNs.}

Physics-informed constraints were integrated either through the model architecture or loss function. Shallow Water Equation–Graph Neural Network (SWE-GNN)~\cite{Bentivoglio2023} modulates propagated messages by the latent gradient between neighboring nodes, which is extended for a multi-scale architecture in multi-scale SWE-GNN (mSWE-GNN)~\cite{Bentivoglio2024}. GNN-fusion~\cite{Zhang2024} computes flow at a node as the sum of predicted flow from its adjacent edges to reliably enforce mass conservation. While it supports node and edge prediction, their model generates these embeddings with different convolutions and has only been tested for 1D flow in drainage networks. HydroGraphNet~\cite{Taghizadeh2025} introduces a physics-based loss that encourages global mass conservation through a forward and backward check, ensuring the predicted water volume is within bounds.

\begin{figure*}[t]
\centering
\includegraphics[width=0.8\textwidth]{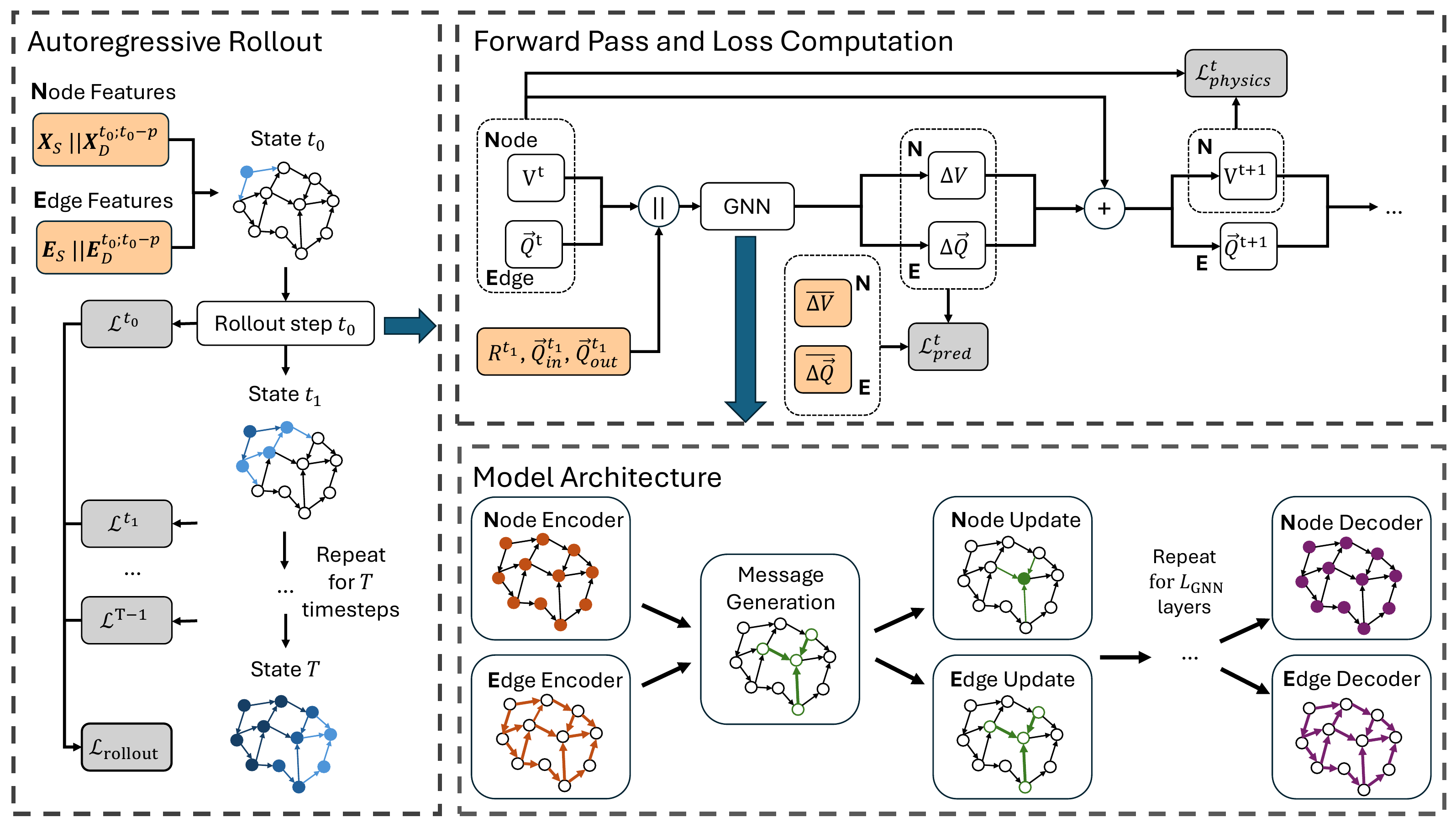}
\caption{The DUALFloodGNN architecture. To predict $T$ timesteps, an autoregressive rollout is performed while accumulating the loss at each timestep to compute the total rollout loss $\mathcal{L}_{rollout}$ (left). During each step, the model predicts node ($\Delta V$) and edge ($\Delta \vec{Q}$) embeddings through joint modeling which shares generated messages (bottom right). From this, the prediction loss $\mathcal{L}_{pred}$ and physics loss $\mathcal{L}_{physics}$ are calculated given the previous state ($V^t$, $\vec{Q}^t$), current state ($V^{t+1} = V^t + \Delta V$, $\vec{Q}^{t+1} = \vec{Q}^t + \Delta \vec{Q}$), and ground truth ($\overline{\Delta V}, \overline{\Delta \vec{Q}}$) (top right). $V$ represents water volume ($m^3$) and $\vec{Q}$ represents water flow ($m^3$/s).}
\label{fig:overview}
\vspace{-3mm}
\end{figure*}

\section{Preliminaries}

\paragraph{Notations.}

Formally, a graph $\mathcal{G} = (\mathcal{V}, \mathcal{E})$ is comprised of a set of nodes $\mathcal{V}$ and edges $\mathcal{E}$. Nodes are described by a node feature matrix $\mathbf{X} = \mathbb{R}^{|\mathcal{V}| \times f_v}$ where $f_v$ is the node feature dimension. Edges are typically represented in terms of an adjacency matrix $\mathbf{A} \in \{0,1\}^{|\mathcal{V}| \times |\mathcal{V}|}$ where $\mathbf{A}_{ij} = 1$ if a connecting edge exists from node $i$ to node $j$, otherwise, $\mathbf{A}_{ij} = 0$. A memory-efficient alternative, especially for sparse graphs, is the coordinate (COO) format $\mathbf{I} = \mathbb{R}^{2 \times |\mathcal{E}|}$ where $(i, j) \in \mathbf{I}$ if there is an edge from node $i$ to node $j$~\cite{Fey2019}. Through this representation, edges can be described with multi-dimensional feature vectors, which combine to form the edge feature matrix $\mathbf{E} = \mathbb{R}^{|\mathcal{E}| \times f_e}$ where $f_e$ is the edge feature dimension.

\paragraph{Problem Definition.}

A flood event is represented as a time series of directed graphs $\{G_0, G_1,...,G_T\}$ where nodes and edges have associated water volume ($V$) and flow ($\vec{Q}$) values respectively. Water flow is a vector quantity where $\vec{Q} > 0$ denotes flow along the edge's orientation and $\vec{Q} < 0$ indicates movement in the opposite direction. Nodes and edges can be further described by static structural properties of the catchment. The model performs a node- and edge-level regression task to predict future hydraulic states $\{V^{t+n}, \vec{Q}^{t+n} | n = 1,...,T\}$ at each structure given the initial state at the current timestep $t$ and up to $p$ previous states $\{V^{t-n}, \vec{Q}^{t-n} | n = 0,...,p\}$. Predictions are obtained in an autoregressive manner, where the predicted features are used recursively as dynamic features for subsequent model inference. Additionally, boundary conditions (e.g., rainfall $R$, upstream inflow $\vec{Q}_{in}$ and downstream outflow $\vec{Q}_{out}$) are supplied at each timestep $\{R^{t+n}, \vec{Q}_{in}^{t+n}, \vec{Q}_{out}^{t+n} | n = 0,...,T-1\}$, describing water flow entering and exiting the system. 

\section{Methodology}
\label{sec:methodology}

\paragraph{Overview.}

In this section, we discuss DUALFloodGNN, a novel physics-informed GNN architecture used for the simultaneous prediction of node water volume and edge water flow. It is comprised of three main components: (1) a model that performs shared message passing to predict both node and edge features, (2) a physics-informed loss function that enforces global and local mass conservation between consecutive predictions, and (3) an autoregressive training strategy utilizing dynamic curriculum learning. Figure~\ref{fig:overview} provides a diagram illustrating this framework.

\subsection{Model Architecture}
\label{sub-sec:architecture}

\paragraph{Input and Output.}

The input of the model is a directed graph at timestep $t$ containing static features (e.g., area, elevation) and dynamic features (e.g., volume, flow, rainfall, global inflow) up to $p$ previous timesteps. The node and edge feature matrices ${\mathbf{X}}^t$ and $\mathbf{E}^t$ are created through Eq. \ref{eq:node-input-feat} and Eq. \ref{eq:edge-input-feat}:
\begin{equation}
    \mathbf{X}^t = \mathbf{X}_S || \mathbf{X}_D^{t-p} || ...||\mathbf{X}_D^t
    \label{eq:node-input-feat}
\end{equation}
\begin{equation}
    \mathbf{E}^t = \mathbf{E}_S || \mathbf{E}_D^{t-p} || ... || \mathbf{E}_D^t
    \label{eq:edge-input-feat}
\end{equation}

\noindent
where $||$ is the concatenation operation. Instead of predicting future states directly, DUALFloodGNN predicts the change in water volume $\Delta V$ and water flow $\Delta \vec{Q}$ from the current timestep to the next timestep. This reduces the variation in the output, achieving more stable and accurate long-horizon predictions. The model's prediction can be added to the current dynamic features $V^t$ and $\vec{Q}^t$ to get the next timestep values $V^{t+1}$ and $\vec{Q}^{t+1}$, which also acts as a residual connection.

To facilitate joint node and edge modeling, we adopt the general Encode-Process-Decode framework~\cite{Battaglia2018}. This is well-suited for the dual regression task as node and edge features are transformed into a unified latent space, enabling a richer exchange of information between them during message passing.

\paragraph{Encoder.}

The encoder block transforms the input node features $\mathbf{X}^t$ and edge features $\mathbf{E}^t$ into their respective latent representations. These two are encoded separately as they represent different physical quantities. The encoding operations are defined in Eqs.~\ref{eq:node_encoder} \&~\ref{eq:edge_encoder} as
\begin{equation}
    \mathbf{H}^t = MLP(\mathbf{X}^t)
    \label{eq:node_encoder}
\end{equation}
\begin{equation}
    \epsilon^t = MLP(\mathbf{E}^t)
    \label{eq:edge_encoder}
\end{equation}

\noindent
where $\mathbf{H}^t = \mathbb{R}^{|\mathcal{V}| \times d}$ and $\mathbf{\epsilon}^t = \mathbb{R}^{|\mathcal{E}| \times d}$ are the encoded node and edge representations, and $d$ is the latent dimension size.

\paragraph{Processor: Shared Message Passing.}

The latent node and edge feature matrices, $\mathbf{H}^t$ and $\epsilon^t$, are then processed through $L_{\text{GNN}}$ message passing layers that iteratively update both of these at each layer $l$. Given a node in the graph $i$, messages $m_{ji}$ are generated for each of its neighboring nodes $\mathcal{N}(i)$:
\begin{equation}
    m_{ji}^{(l)} = MLP(h_i^{(l)} || h_j^{(l)} || e_{ij}^{(l)}); \: \forall j \in \mathcal{N}(i)
\end{equation}

\noindent
where $h_{i}$ and $h_{j}$ are the node embeddings and its neighbor embeddings, respectively, and $e_{ij}$ is the edge embedding between them. Messages are conditioned on both node and edge attributes, as they are high-dimensional representations of flood volume along their respective structures, suggesting their co-dependency in characterizing local fluid dynamics.

Following this interpretation, the same generated messages are used to simultaneously update both embedding types. To update the current embedding of node $i$, messages $m_{ji}$ are aggregated through summation, improving the model's expressive power in identifying the underlying graph structure~\cite{Xu2019}. The aggregated value is fed to a parametrized fully connected layer to generate the updated node embeddings $h^{(l+1)}$:
\begin{equation}
    h^{(l+1)}_i = MLP(\sum_{j \in \mathcal{N}(i)} m_{ji}^{(l)})
\end{equation}

\noindent
The updated edge embeddings $e_{ij}^{(l+1)}$ are directly overwritten with the message $m_{ji}$ at each edge, inspired by the approach of Ashraf et al.~\shortcite{Ashraf2024}.
\begin{equation}
    e_{ij}^{(l+1)} = m_{ji}^{(l)}; \: \forall j \in \mathcal{N}(i)
\end{equation}

\noindent
Residual connections are added at each node and edge update to stabilize predictions. From a hydrological perspective, these update operations are physically meaningful, as messages $m_{ji}$ symbolize flow between regions and their sum $\Sigma_{j \in \mathcal{N}(i)} m_{ji}$ for node $i$ would represent the net water flux in that region.

\paragraph{Decoder.}

After the processor block, the final embedding matrices of the nodes and edges, $\mathbf{H}^L$ and $\epsilon^L$, are translated back to the prediction feature space through a decoder, generating the final output of the model. Similar to the encoder, two separate modules are used to produce predictions for each type of structure. This decoding operation can be summarized by Eqs.~\ref{eq:node_decoder} \&~\ref{eq:edge_decoder} as
\begin{equation}
    \Delta V^{t+1} = MLP(\mathbf{H}^L)
    \label{eq:node_decoder}
\end{equation}
\begin{equation}
    \Delta \vec{Q}^{t+1} = MLP(\mathbf{\epsilon}^L)
    \label{eq:edge_decoder}
\end{equation}

\noindent
where $\Delta V^{t+1} = \mathbb{R}^{|\mathcal{V}|}$ and $\Delta \vec{Q}^{t+1} = \mathbb{R}^{|\mathcal{E}|}$ are the respective node and edge predictions.

\paragraph{Learnable Parameters.}

The encoders, decoders and message passing functions are all implemented as MLPs with $L_{\text{MLP}}$ layers. A ReLU activation function is added after each layer $l$ except for the final layer $L_{\text{MLP}}-1$. This is mathematically expressed in Eqs.~\ref{eq:mlp_update} \&~\ref{eq:mlp_update_last_layer}:
\begin{equation}
    \mathbf{x}^{l+1}= ReLU(\mathbf{W}^{l}\mathbf{x}^{l})
    \label{eq:mlp_update}
\end{equation}
\begin{equation}
    \mathbf{x}^{(L_{\text{MLP}})}= \mathbf{W}^{(L_{\text{MLP}}-1)}\mathbf{x}^{(L_{\text{MLP}}-1)}
    \label{eq:mlp_update_last_layer}
\end{equation}

\noindent
where $\mathbf{x}$ are the processed embeddings and $\mathbf{W}$ is a matrix with learnable parameters. Following the design of SWE-GNN~\cite{Bentivoglio2023}, the bias term is omitted to prevent spurious water volume predictions in isolated nodes distant from the flood extent, a modification empirically demonstrated to improve flood classification.

\subsection{Physics-Informed Loss Function}
\label{sub-sec:physics-loss}

\paragraph{Multi-task Prediction Loss.}

The base training objective is formulated as a multi-task loss consisting of the mean squared error of node and edge predictions, as defined in Eqs.~\ref{eq:node-mse} \&~\ref{eq:edge-mse}.
\begin{equation}
    \mathcal{L}_{node} = \frac{1}{|\mathcal{V}|} \sum_{i \in \mathcal{V}} (\overline{\Delta V_i} - \Delta V_i)^2
    \label{eq:node-mse}
\end{equation}
\begin{equation}
    \mathcal{L}_{edge} = \frac{1}{|\mathcal{E}|} \sum_{k \in \mathcal{E}} (\overline{\Delta \vec{Q}_{k}} - \Delta \vec{Q}_{k})^2.
    \label{eq:edge-mse}
\end{equation}

The total prediction loss can then be computed as a weighted sum between these two terms, expressed in Eq. \ref{eq:prediction-loss}:
\begin{equation}
    \mathcal{L}_{pred} = \lambda_1 \mathcal{L}_{node} + \lambda_2 \mathcal{L}_{edge}
    \label{eq:prediction-loss}
\end{equation}
where the coefficients $\lambda_1$ and $\lambda_2$ dictate the contribution of $\mathcal{L}_{node}$ and $\mathcal{L}_{edge}$ in $\mathcal{L}_{pred}$.

\paragraph{Mass Balance Regularization.}

To ground model predictions in physical realism, we utilize additional regularization terms derived from the conservation of mass from the Shallow Water Equations, the fundamental principles governing 2D flood movement ~\cite{Costabile2017}. If assuming constant density, mass conservation is equivalent to volume conservation, enabling formulation using volumetric quantities aligned with the model's predictions. Given a control volume $CV$, this can be expressed as
\begin{equation}
    \frac{dV}{dt} = \sum_{CV} \vec{Q}_{+} - \sum_{CV} \vec{Q}_{-} + \vec{Q}_{\text{sources}} - \vec{Q}_{\text{sinks}}
\end{equation}

\noindent
where $\vec{Q}_{+}$ and $\vec{Q}_{-}$ is the influx and efflux through $CV$, and $\vec{Q}_{\text{sources}}$ and $\vec{Q}_{\text{sinks}}$ are flows adding and removing water from the system. The physics-informed loss is comprised of global mass loss $\mathcal{L}_{global}$ and local mass loss $\mathcal{L}_{local}$, which is written in Eq. \ref{eq:physics-loss} as
\begin{equation}
    \mathcal{L}_{physics} = \lambda_3 \mathcal{L}_{global} + \lambda_4 \mathcal{L}_{local}
    \label{eq:physics-loss}
\end{equation}

\noindent
where the coefficients $\lambda_3$ and $\lambda_4$ balance the contribution of $\mathcal{L}_{global}$ and $\mathcal{L}_{local}$ in $\mathcal{L}_{physics}$. We now explain the formulation of $\mathcal{L}_{global}$ and $\mathcal{L}_{local}$.

\paragraph{Global-level Regularization.}

Global mass conservation considers the entire target catchment as the control volume $CV$. We define the global mass conservation loss as follows (Eq. \ref{eq:global-mass-loss}):
\begin{equation}
    \label{eq:global-mass-loss}
    \mathcal{L}_{global} = |\sum_{i \in \mathcal{V}} \Delta V_{i}^t - ((\vec{Q}_{in}^t - \vec{Q}_{out}^t) \cdot \Delta t + \sum_{i \in \mathcal{V}} R_{i}^t)|
\end{equation}

\noindent
where $\Delta V_i^t = V_{i}^{t+1} - V_{i}^t$ is the change in volume at node $i$ from the current timestep $t$ to the next timestep $t+1$ and $\Delta t$ is the temporal resolution between consecutive timesteps. The regularization term is based on the absolute value of the mass balance residual, resulting in a well-defined convex objective ($\mathcal{L}_{global} = 0$) for gradient-based optimization. Notably, this form of regularization is only applied to node volume predictions $V$ as the other variables $\vec{Q}_{in}$, $\vec{Q}_{out}$ and $R$ are given to the model as boundary conditions.

\paragraph{Local-level Regularization.}

Mass conservation was also explicitly enforced on a local scale considering each individual node $i$ as the control volume $CV$. We define the local mass conservation loss as follows (Eq. \ref{eq:local-mass-loss}):
\begin{equation}
    \label{eq:local-mass-loss}
    \mathcal{L}_{local} = \sum_{i \in \mathcal{V}} |\Delta V_{i}^t - ((\vec{Q}_{i+}^t - \vec{Q}_{i-}^t) \cdot \Delta t + R_{i}^t)|
\end{equation}

\noindent
where $\vec{Q}_{i+}$ and $\vec{Q}_{i-}$ are the respective inflow and outflow at a node $i$. Taking the absolute value is performed for the same reasons previously mentioned, where $\mathcal{L}_{local} = 0$ corresponds to perfect conservation at all nodes. Unlike the global regularization term, this condition balances both node water volume $V$ and edge water flow $\vec{Q}$ predictions.

The inflow $\vec{Q}_{i+}$ and outflow $\vec{Q}_{i-}$ at each node are computed directly from the edge predictions $\vec{Q}$. Given the 
COO adjacency matrix $\mathbf{I}$, let $\mathbf{I}' = \mathbb{R}^{2 \times |\mathcal{E}|}$ be the transposed adjacency matrix such that $\mathbf{I}' = \{ (j, i) \mid (i, j) \in \mathbf{I} \}$. The undirected adjacency matrix $\mathbf{I}_{\text{undir}} = \mathbb{R}^{2 \times 2|\mathcal{E}|}$ can then be represented as in Eq. \ref{eq:undir_adj_matrix}.
\begin{equation}
    \mathbf{I}_{\text{undir}} = \mathbf{I} \: || \: \mathbf{I}'.
    \label{eq:undir_adj_matrix}
\end{equation}

\noindent
Weights are generated for each edge in $\mathbf{I}_{\text{undir}}$ representing the flow along its direction through Eq. \ref{eq:weighted_adj_matrix}.
\begin{equation}
    \mathbf{I}_w = \mathbf{I}_{\text{undir}} \odot ReLU(\vec{Q} || -\vec{Q})
    \label{eq:weighted_adj_matrix}
\end{equation}

\noindent
where $\odot$ is the element-wise product. Applying $ReLU(\cdot)$ removes negative values of $\vec{Q}$, which indicate flow opposite to the edge direction, leaving only positive weights. The total inflow is then computed for each node $i$ as in Eq. \ref{eq:total_node_inflow}:
\begin{equation}
    \vec{Q}_{i+} = \sum_{j \in \mathcal{N}_{\text{in}}(i)} \mathbf{I}_{w_{ij}}
    \label{eq:total_node_inflow}
\end{equation}

\noindent
where $\mathcal{N}_{\text{in}}(i)$ are the edges pointing towards node $i$. Similarly, the total outflow at each node is denoted as in Eq. \ref{eq:total_node_outflow}:
\begin{equation}
    \vec{Q}_{i-} = \sum_{j \in \mathcal{N}_{\text{out}}(i)} \mathbf{I}_{w_{ji}}
    \label{eq:total_node_outflow}
\end{equation}

\noindent
where $\mathcal{N}_{\text{out}}(i)$ are the edges pointing away from node $i$.

\paragraph{Overall Loss.}

The overall loss function is derived by taking the sum of the prediction loss and physics loss, defined in Eq. \ref{eq:overall-loss}:
\begin{equation}
    \mathcal{L} = \mathcal{L}_{pred} + \mathcal{L}_{physics}.
    \label{eq:overall-loss}
\end{equation}

\subsection{Autoregressive Training}

\paragraph{Multi-step-ahead Loss.}

Multi-step predictions are obtained through autoregressive inference that introduces gradually accumulating errors. The typical supervised or teacher forcing training strategy is unable to handle such distribution shifts~\cite{Teutsch2022}. We adopt the approach of Bentivoglio et al.~\shortcite{Bentivoglio2023}, which  proposes the use of a multi-step-ahead loss to allow the model to develop robustness to its own noisy input. For each training batch, an autoregressive rollout is performed, after which the total loss used for backpropagation is calculated as in Eq. \ref{eq:multi-step-loss}:
\begin{equation}
    \label{eq:multi-step-loss}
    \mathcal{L}_{rollout} = \frac{1}{O} \sum^{O}_{o=1} \mathcal{L}(\Delta V^o, \Delta \vec{Q}^o)
\end{equation}

\noindent
where $O$ is the rollout length and $\mathcal{L}$ is the overall loss defined in Eq.~\ref{eq:overall-loss}. This procedure is particularly crucial for the application of the physics-informed loss, as during multi-step rollouts, the constrained hydrological states are given purely by the model's predictions. Thus, not only does the model learn to correct its own errors, but to do so in a manner that adheres to mass conservation throughout the trajectory.

\paragraph{Dynamic Curriculum Learning.}

To improve training stability, the original implementation utilizes a curriculum learning strategy~\cite{Bentivoglio2023}. The model is initially trained on the simpler task of predicting 1-step-ahead loss, which is gradually increased at a 1-step increment after a set number of epochs until it finally reaches the target rollout length $O$. However, this deterministic approach fails to accommodate the non-uniform difficulty of predicting varying time horizons. To create a more adaptive progression, this fixed schedule is replaced with an early stopping mechanism, where the prediction horizon is extended only after the model's performance has converged on the validation set for the current rollout length $o$. Furthermore, we design the step size between curriculum stages to be a tunable hyperparameter $C$, which controls the learning progression. To mitigate catastrophic interference, the learning rate is decayed by a factor $\gamma$ after each curriculum step. This entire process can be interpreted as a sequence of fine-tuning steps, where a converged model for a given horizon is adapted for the subsequent, more challenging longer rollout duration.

\section{Experiments}

\subsection{Experimental Setup}

\paragraph{Dataset.}

The target catchment for this study was taken from a section of the Wollombi River watershed located in New South Wales, Australia. A dataset comprised of 56 flow-dominant flood events was generated using the HEC-RAS numerical model~\cite{Brunner2025}. The 2D unstructured mesh was converted to a directed graph composed of 1129 nodes and 2743 edges. Each event consists of 576 time steps with a resolution of 15 minutes.

\paragraph{Baselines.}

The DUALFloodGNN architecture was compared to standard GNN models such as Graph Convolutional Network (GCN)~\cite{Kipf2017},
Graph Attention Network (GAT)~\cite{Velickovic2018},
Graph SAmple and aggreGatE (GraphSAGE)~\cite{Hamilton2018}, Graph Isomorphism Network (GIN)~\cite{Xu2019}, and Graph Isomorphism Network
with Edges (GINE)~\cite{Hu2020}. Variations of these models for edge regression were created by incorporating an MLP prediction head that leverages the features of adjacent nodes to predict edge values. The proposed architecture was also compared to domain-specific GNN models, including HydroGraphNet~\cite{Taghizadeh2025} and SWE-GNN~\cite{Bentivoglio2023}. Models were evaluated for the water volume, flow and depth prediction tasks through 14-fold cross validation.

\paragraph{Metrics.}

Performance for the regression tasks was measured using Root Mean Squared Error (RMSE) and Mean Absolute Error (MAE) per timestep. To measure flood extent classification, Critical Success Index (CSI), which is defined in Eq. \ref{eq:csi_formula}, was computed by classifying flooded nodes for the thresholds $\tau = 0.05 \ m$ and $\tau = 0.3 \ m$, following the convention in other literature~\cite{Bentivoglio2023,Herath2025,Taghizadeh2025}.
\begin{equation}
    \text{CSI} = \frac{TP}{TP + FN + FP}.
    \label{eq:csi_formula}
\end{equation}


\begin{table}[t]
\centering
\scriptsize
\begin{tabularx}{\columnwidth}{XXX}
\toprule
\textbf{Model} & \textbf{RMSE} {\tiny ($m^3$)} $\downarrow$ & \textbf{MAE} {\tiny ($m^3$)} $\downarrow$ \\
\midrule
GCN (node) & 9,849.98 {\tiny $\pm$ 2,485.73} & 6,340.51 {\tiny $\pm$ 1,859.60} \\
GAT (node) & 13,232.86 {\tiny $\pm$ 11,835.18} & 4,583.68 {\tiny $\pm$ 1,969.27} \\
GraphSAGE (node) & 11,146.97 {\tiny $\pm$ 11,420.79} & 4,177.75 {\tiny $\pm$ 2,519.14} \\
GIN (node) & 5,023.18 {\tiny $\pm$ 1,450.37} & 2,587.05 {\tiny $\pm$ 798.38} \\
GINE (node) & \underline{3,437.68 {\tiny $\pm$ 693.94}} & \underline{1,860.03 {\tiny $\pm$ 406.78}} \\
HydroGraphNet & 7,584.51 {\tiny $\pm$ 3,068.54} & 3,343.52 {\tiny $\pm$ 1,392.34} \\
DUALFloodGNN & \textbf{2,216.27} {\tiny $\pm$ \textbf{868.21}} & \textbf{1,075.63} {\tiny $\pm$ \textbf{394.83}} \\
\bottomrule
\end{tabularx}
\caption{Average metrics ($\pm$ std) for the node water volume regression task. The best performing model is highlighted in bold while the second best result is underlined.}
\label{tab:vol-results}
\end{table}

\subsection{Experimental Results}

\paragraph{Overall Results.}

The results for the water volume and flow prediction task are shown in tables~\ref{tab:vol-results} and ~\ref{tab:flow-results} respectively. For node volume prediction, DUALFloodGNN yielded the best results out of all the models, surpassing the RMSE of the second best model, GINE, by $35.53 \%$. The proposed model maintained its superior standing for edge flow regression compared to other benchmarked GNNs, improving the baseline RMSE by $41.21 \%$. Notably, baseline models that incorporated edge attributes (GAT and GINE) performed significantly worse in flow prediction, highlighting the importance of joint node and edge modeling for utilizing edge features. In terms of variability, DUALFloodGNN exhibited stable performance across different test sets, yielding a low standard deviation. Moreover, only DUALFloodGNN obtained metrics for both water volume and flow within a single model whereas other baseline architectures required multiple models for each prediction task.

\begin{table}[t]
\centering
\scriptsize
\begin{tabularx}{\columnwidth}{XXX}
\toprule
\textbf{Model} & \textbf{RMSE} {\tiny ($m^3/s$)} $\downarrow$ & \textbf{MAE} {\tiny ($m^3/s$)} $\downarrow$ \\
\midrule
GCN (edge) & 90.99 {\tiny $\pm$ 25.04} & 40.37 {\tiny $\pm$ 10.47} \\
GAT (edge) & 263.40 {\tiny $\pm$ 63.16} & 60.11 {\tiny $\pm$ 13.90} \\
GraphSAGE (edge) & \underline{44.02 {\tiny $\pm$ 12.54}} & \underline{17.05 {\tiny $\pm$ 4.48}} \\
GIN (edge) & 51.21 {\tiny $\pm$ 16.33} & 22.64 {\tiny $\pm$ 7.61} \\
GINE (edge) & 249.31 {\tiny $\pm$ 70.59} & 58.38 {\tiny $\pm$ 15.56} \\
DUALFloodGNN & \textbf{25.88} {\tiny $\pm$ \textbf{9.64}} & \textbf{11.40} {\tiny $\pm$ \textbf{4.17}} \\
\bottomrule
\end{tabularx}
\caption{Average metrics ($\pm$ std) for the edge water flow regression task. The best performing model is highlighted in bold while the second best result is underlined.}
\label{tab:flow-results}
\end{table}

\begin{table}[t]
\centering
\scriptsize
\setlength{\tabcolsep}{2.5pt}
\begin{tabularx}{\columnwidth}{lXXXXX}
\toprule
\multirow{2}{*}{\textbf{Model}} & \textbf{RMSE} $\downarrow$ & \textbf{MAE} $\downarrow$ & \multicolumn{2}{c}{\textbf{CSI} $\uparrow$} & \multirow{2}{=}{\textbf{Inference Time} ($s$)} \\
\cline{4-5}
& ($m$) & ($m$) & $\tau$ = 0.05 $m$ & $\tau$ = 0.3 $m$ & \\
\midrule
HydroGraphNet & \underline{0.76 {\tiny $\pm$ 0.34}} & 0.29 {\tiny $\pm$ 0.16} & \underline{0.69 {\tiny $\pm$ 0.13}} & \underline{0.80 {\tiny $\pm$ 0.05}} & \textbf{3.25} {\tiny $\pm$ \textbf{0.17}} \\
SWE-GNN & 1.23 {\tiny $\pm$ 0.35} & \underline{0.61 {\tiny $\pm$ 0.21}} & 0.47 {\tiny $\pm$ 0.11} & 0.49 {\tiny $\pm$ 0.09} & 7.21 {\tiny $\pm$ 0.20} \\
DUALFloodGNN & \textbf{0.21} {\tiny $\pm$ \textbf{0.07}} & \textbf{0.07} {\tiny $\pm$ \textbf{0.03}} & \textbf{0.90} {\tiny $\pm$ \textbf{0.03}} & \textbf{0.91} {\tiny $\pm$ \textbf{0.03}} & \underline{4.10 {\tiny $\pm$ 0.06}} \\
\bottomrule
\end{tabularx}
\caption{Average metrics ($\pm$ std) for the node water depth regression task. The best performing model is highlighted in bold while the second best result is underlined.}
\label{tab:wd-results}
\end{table}

\begin{figure*}[t]
    \centering
    \begin{subfigure}{0.3\textwidth}
        \centering
        \includegraphics[width=\textwidth]{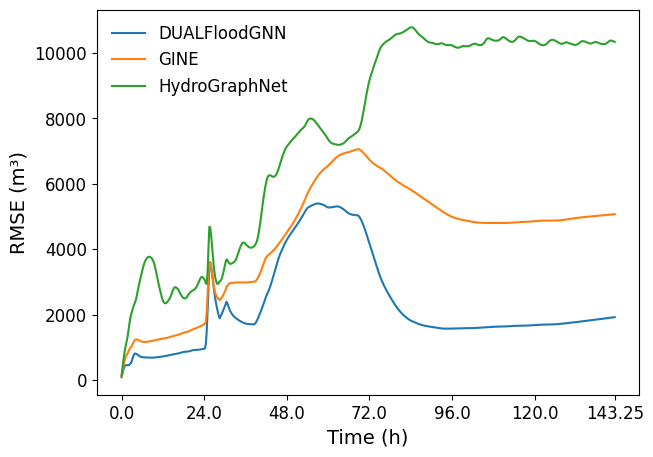}
        \caption{Water volume ($m^3$)}
    \end{subfigure}
    \hspace{0.25cm}
    \begin{subfigure}{0.3\textwidth}
        \centering
        \includegraphics[width=\textwidth]{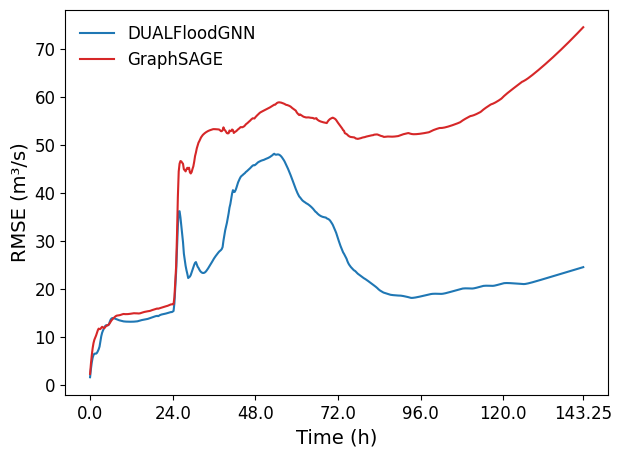}
        \caption{Water flow ($m^3 / s$)}
    \end{subfigure}
    \hspace{0.25cm}
    \begin{subfigure}{0.3\textwidth}
        \centering
        \includegraphics[width=\textwidth]{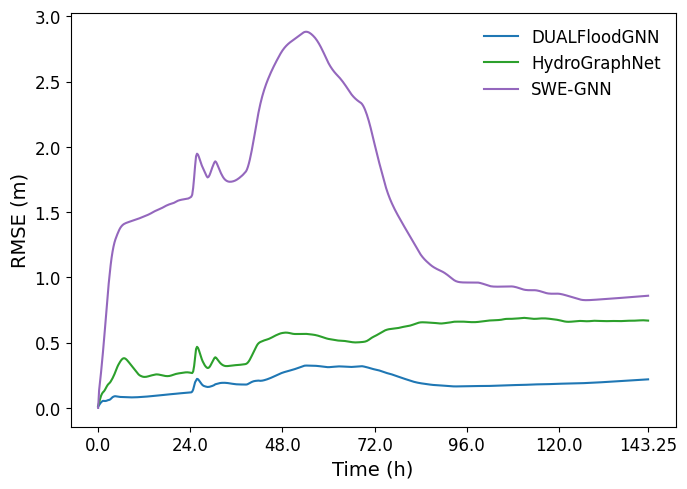}
        \caption{Water depth ($m$)}
    \end{subfigure}
    \caption{Task-specific RMSE of select benchmarked GNN models for each timestep in the case study event.}
    \label{fig:event_rmse}
\end{figure*}

Each flood GNN model's performance was evaluated for water depth prediction given its prevalence in practical use-cases. Water depth was back calculated from DUALFloodGNN's predictions with a volume-to-elevation mapping provided in the HEC-RAS simulation. As observed in table~\ref{tab:wd-results}, the proposed model continued to outperform even domain-specific models, yielding a $72.37 \%$ improvement to the second best RMSE. In terms of classification, DUALFloodGNN accurately categorized flooded nodes by achieving a similar CSI of around 0.9 in both 0.05 and 0.3 m thresholds. SWE-GNN showed the same trend albeit at a much lower accuracy, while HydroGraphNet displayed a high delta between its CSI for the two cases. With regards to inference speed, DUALFloodGNN closely matches the low latency of HydroGraphNet with only a 0.85 second difference. Thus, DUALFLoodGNN preserves the same 2 to 3 orders of magnitude speed-up compared to numerical solvers as reported by the previous studies~\cite{Bentivoglio2023,Taghizadeh2025}. For example, an event that would take almost a day to run for a numerical model would only take around 3 minutes for our model to simulate.

\begin{figure}[t]
\centering
\includegraphics[width=0.85\columnwidth]{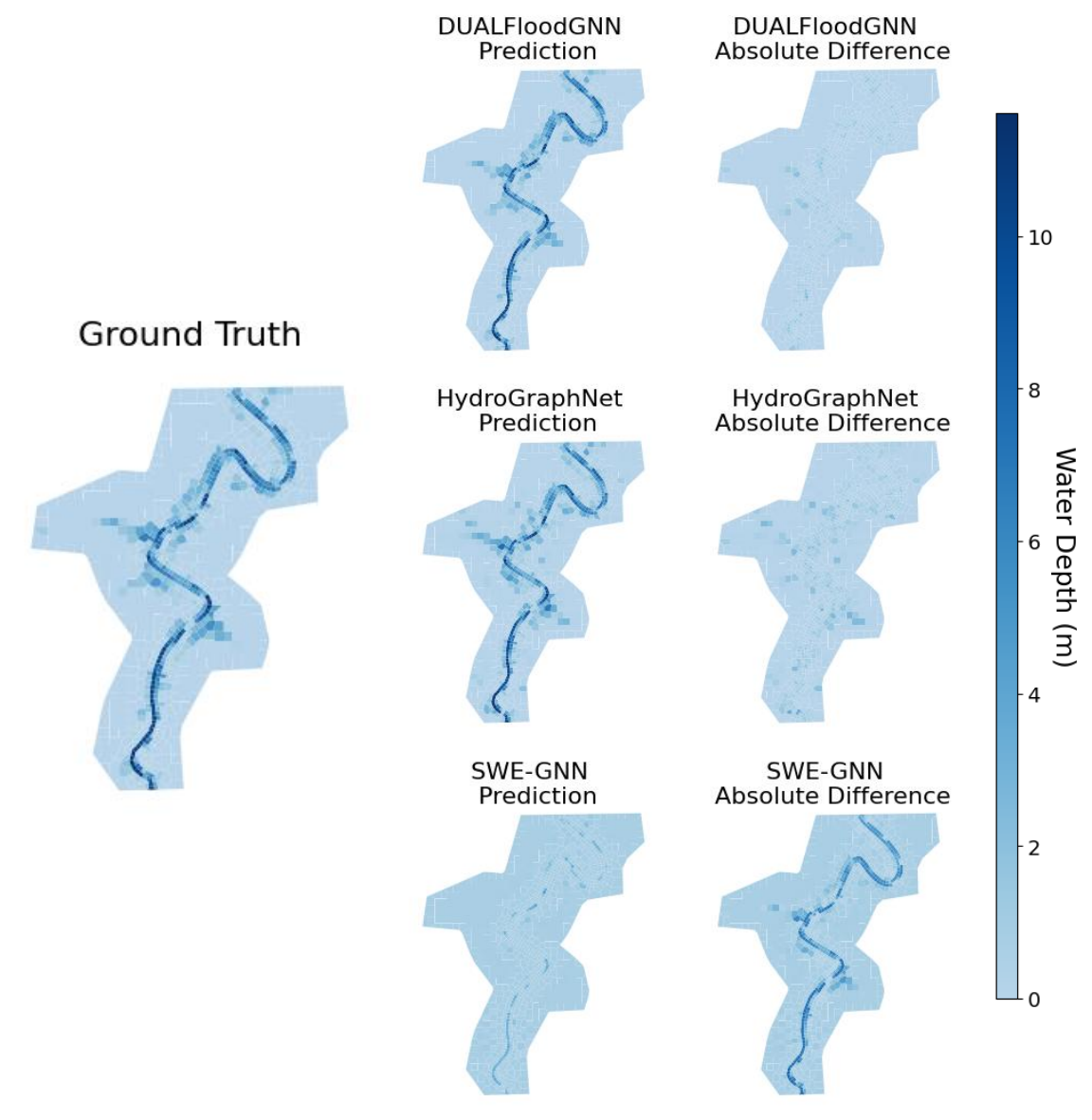}
\caption{Maximum water depth map of flood-specific GNN models for the case study event. The maps of the ground truth, model predictions (left) and difference between the two (right) are shown.}
\label{fig:event_max_depth_map}
\vspace{-3mm}
\end{figure}

\paragraph{Event Analysis.}

The water volume, flow and depth RMSE per timestep of select GNNs were plotted for a single case study event in figure~\ref{fig:event_rmse}. Across all tasks, DUALFloodGNN exhibited superior temporal stability, consistently maintaining the lowest RMSE values throughout the entire rollout period. To visualize flood extent, max flood depth maps, which plot the maximum water depth for each node in the graph across all timesteps, were generated for the selected event (figure ~\ref{fig:event_max_depth_map}). In the ground truth, the nodes with the highest water depth are located near the river area where flow is most concentrated. SWE-GNN predicted mostly uniform flood propagation throughout the catchment while HydroGraphNet overestimated flooding for a few nodes far from the central river area. DUALFloodGNN produced the most accurate map, demonstrating its capacity to capture the flood's microscale dynamics.

\subsection{Ablation Studies}
\label{sub-sec:ablation}

Table \ref{tab:ablation} summarizes the ablation studies performed to identify the key factors contributing to the model's performance.

\begin{table}[t]
\centering
\scriptsize
\begin{tabularx}{\columnwidth}{XXX}
\toprule
\multirow{2}{\linewidth}{\textbf{Model}} & \textbf{Node RMSE} $\downarrow$ & \textbf{Edge RMSE} $\downarrow$ \\
& ($m^3$) & ($m^3/s$) \\
\midrule
DUALFloodGNN & 2,216.27 {\tiny $\pm$ 868.21} & \textbf{25.88} {\tiny $\pm$ \textbf{9.64}} \\
\hspace{0.3cm} No Inflow Feature & 20,562.71 {\tiny $\pm$ 8,054.11} & 423.40 {\tiny $\pm$ 246.11} \\
\hspace{0.3cm} Global Loss Only & 2,794.10 {\tiny $\pm$ 851.61}  & 48.09 {\tiny $\pm$ 14.52} \\
\hspace{0.3cm} Local Loss Only & \textbf{2,116.45} {\tiny $\pm$ \textbf{845.44}}  & 26.07 {\tiny $\pm$ 8.55} \\
\hspace{0.3cm} No Physics Loss & 2,640.42 {\tiny $\pm$ 1011.80}  & 46.86 {\tiny $\pm$ 15.81} \\
\midrule
HydroGraphNet & \textbf{7,584.51} {\tiny $\pm$ \textbf{3,068.54}} & - \\
\hspace{0.3cm} No Inflow Feature & 20,867.39 {\tiny $\pm$ 6697.08} & - \\
\bottomrule
\end{tabularx}
\caption{Ablation study comparing RMSE across different model configurations. The best metrics are highlighted in bold.}
\label{tab:ablation}
\vspace{-3mm}
\end{table}

\paragraph{Global Inflow Feature.}

The exclusion of the global inflow $Q_{in}$ significantly affected the predictive accuracy of the model. The inclusion of this global feature~\cite{Battaglia2018} for all nodes is an approach directly adopted from HydroGraphNet~\cite{Taghizadeh2025}, which demonstrated the same degradation in performance when omitted. This could be attributed to the flow-dominant nature of the flood events in the dataset. We enhanced the global inflow feature by treating it as a dynamic feature, also including the previous boundary conditions as model input.

\paragraph{Physics-informed Loss}

Three additional physics-informed variants were considered: only global loss, only local loss and no physics-informed loss. In contrast to applying only global mass balance, integrating local-level regularization resulted in a lower RMSE for both node and edge prediction, surpassing the accuracy of the purely data-driven model. This can be attributed to the co-dependent modeling of node and edge outputs in the local physics loss. A model with physics constraints on both scales demonstrates comparable overall performance to one with only local loss, though the superior variant varies across testing folds. Thus, selection between the two should be driven by empirical evaluation on the target dataset.

\section{Conclusion}
\label{sec:conclusion}

In conclusion, this study presents DUALFloodGNN, a novel physics-informed flood GNN, that explicitly incorporates global and local mass conservation within the loss function. To this end, it performs simultaneous node and edge prediction, jointly modeling these values through a shared message passing operation. To enhance robustness, the model is trained with a dynamic curriculum learning strategy, which adjusts the learning period in response to the current difficulty. DUALFloodGNN outperformed standard GNN architectures and state-of-the-art GNN flood models in the prediction of hydrodynamic variables and the classification of flooded locations. Additionally, the model maintained computational efficiency comparable to other flood-specific GNN approaches, demonstrating its suitability for operational use cases.

Future research can evaluate the model's generalizability for other catchments and flood types. For example, in flows with high sediment or debris concentration, density is not spatially uniform which may affect volume-based regularization.

\bibliographystyle{named}
\bibliography{references}

\end{document}